\documentclass[10pt,twocolumn,letterpaper]{article}

\usepackage[pagenumbers]{cvpr} 

\usepackage{graphicx}
\usepackage{amsmath}
\usepackage{graphbox}
\DeclareMathOperator{\keeptop}{TOP1}
\DeclareMathOperator{\softmax}{softmax}
\usepackage{xr}
\usepackage{amssymb}
\usepackage{booktabs}
\usepackage[none]{hyphenat}

\usepackage[pagebackref,breaklinks,colorlinks]{hyperref}

\usepackage[capitalize]{cleveref}
\crefname{section}{Sec.}{Secs.}
\Crefname{section}{Section}{Sections}
\Crefname{table}{Table}{Tables}
\crefname{table}{Tab.}{Tabs.}

\def\cvprPaperID{4861} 
\def\confName{CVPR}
\def\confYear{2023}

\begin{document}

\title{Efficient Zero-shot Visual Search via Target and Context-aware Transformer}


\author {
    Zhiwei Ding\textsuperscript{\rm 1,2,*},
    Xuezhe Ren\textsuperscript{\rm 3,*},
    Erwan David\textsuperscript{\rm 4},
    Melissa Vo\textsuperscript{\rm 4},
    Gabriel Kreiman\textsuperscript{\rm 5,6},
    Mengmi Zhang\textsuperscript{\rm 3} \\
    \textsuperscript{\rm *} \small Equal contribution\\
    \textsuperscript{\rm 1} \small Department of Neuroscience, Baylor College of Medicine, Houston, TX, USA, \\
    \textsuperscript{\rm 2} \small Center for Neuroscience and Artificial Intelligence, Baylor College of Medicine, Houston, TX, USA, \\
    \textsuperscript{\rm 3} \small CFAR and I2R, Agency for Science, Technology and Research, Singapore, \\
    \textsuperscript{\rm 4} \small Department of Psychology, Goethe Univerisity Frankfurt, Frankfurt, Germany, \\
    \textsuperscript{\rm 5} \small Children's Hospital, Harvard Medical School, Boston, MA, USA, \\
    \textsuperscript{\rm 6} \small Center for Brains, Minds and Machines, Cambridge, MA, USA, \\
    \small Address correspondence to mengmi@i2r.a-star.edu.sg
}

\maketitle

\begin{abstract}
Visual search is a ubiquitous challenge in natural vision, including daily tasks such as finding a friend in a crowd or searching for a car in a parking lot. Human rely heavily on relevant target features to perform goal-directed visual search. Meanwhile, context is of critical importance for locating a target object in complex scenes as it helps narrow down the search area and makes the search process more efficient. However, few works have combined both target and context information in visual search computational models. Here we propose a zero-shot deep learning architecture, \textbf{TCT} (Target and Context-aware Transformer), that modulates self attention in the Vision Transformer with target and contextual relevant information to enable human-like zero-shot visual search performance. Target modulation is computed as patch-wise local relevance between the target and search images, whereas contextual modulation is applied in a global fashion. We conduct visual search experiments on TCT and other competitive visual search models on three natural scene datasets with varying levels of difficulty. TCT demonstrates human-like performance in terms of search efficiency and beats the SOTA models in challenging visual search tasks. Importantly, TCT generalizes well across datasets with novel objects without retraining or fine-tuning. Furthermore, we also introduce a new dataset to benchmark models for invariant visual search under incongruent contexts. TCT manages to search flexibly via target and context modulation, even under incongruent contexts.   



   

\end{abstract}




\section{Introduction} \label{sec:intro}
\begin{figure}
\centering
\includegraphics[scale=0.35]{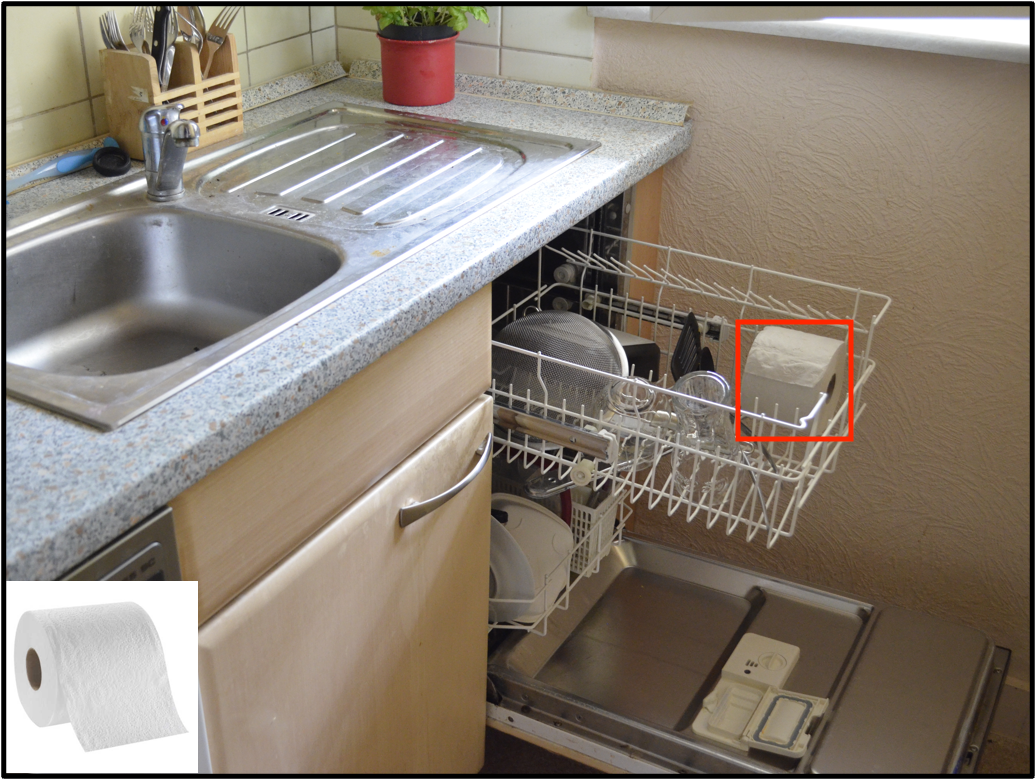}
\caption{\textbf{Searching for targets in semantically incongruent scenes is inefficient.} The image describes a kitchen scene photograph with a red bounding box indicating the location of target object (toilet paper, shown also in the bottom left corner). The image has a semantic object-scene violation: the target object does not fit into the global meaning of the scene (i.e. toilet paper inside a dishwasher in kitchen). 
}\vspace{-4mm}
\label{fig:teaser}
\end{figure}

When 
searching for toilet paper, we usually expect a bathroom scene and will likely search through regions around the toilet or the sink for a white and short cylinder-shaped object. If provided with a kitchen scene that is irrelevant to our object of interest, we might feel lost about where to focus on and simply search randomly throughout the scene; nonetheless, our expectation of toilet paper's visual appearance will still lead us to find it without too much difficulty (\textbf{Fig.~\ref{fig:teaser}}). Humans rely heavily on knowledge about the target object and the surrounding context in daily visual search tasks. Target information provides reference of specific visual features to guide attention during the search, whereas context information helps narrow down the search area and guide attention towards more relevant locations. Both sources of information work coherently to make visual search more efficient. 

Visual search constitutes a ubiquitous challenge in daily life and localizing a target object in a complex scene is important in many applications. Visual search must fulfill four key properties: (1) selectivity (to distinguish the target from distractors in a cluttered scene), (2) invariance (to localize the target despite differences in its appearance or in cases where the target is partially occluded), (3) efficiency (to localize the target with a few eye movements instead of exhaustive sampling) and (4) zero-shot (to generalize to novel target objects despite minimal or zero prior exposure to them). Given a search image and a target object, it remains unclear how to build a computational model that capitalizes on both target and contextual cues to efficiently localize the target object and generalize to various search conditions without additional training. Furthermore, investigation on the impact of incongruent context on invariant visual search performance is limited, largely due to the lack of such dataset. Synthetic out-of-context images have been introduced previously \cite{bomatter2021pigs}, but those images are not realistic.

Here we propose a zero-shot deep learning architecture, the Target and Context-aware Transformer (\textbf{TCT}), that integrates both target and contextual information to modulate self-attention via a multi-head transformer encoder as in the Vision Transformer (ViT) \cite{dosovitskiy2020image}. Target modulation is computed as the patch-wise local relevance between the target and search images, whereas contextual modulation is applied in a global fashion. Unlike the existing visual search models that learn to predict fixation sequences based on human eye tracking data \cite{yang2020predicting}, our model is zero-shot where it directly transfers knowledge from contextual reasoning in object recognition tasks to visual search. 

We test our model on three image datasets spanning various levels of difficulty. TCT beats the SOTA models in highly difficult datasets, handling complex scenes with clutter distractors and incongruent contexts without task-specific fine-tuning. With built-in target and context modulation, TCT can flexibly use both sources of information to strategize for high search efficiency. 

In summary, our key contributions include:

\noindent \textbf{[model]} We propose a biologically inspired search model with a hierarchical deep architecture based on the pre-trained ViT. Without additional training, the model can efficiently find novel target objects in cluttered natural scenes, despite changes in the target's color, scale, rotation or even when the search scene contains a different exemplar belonging to the same category as the target. 

\noindent \textbf{[dataset]} We introduce a new image dataset to benchmark invariant visual search under natural incongruent contexts.

\noindent \textbf{[zero-shot and human-like]} Our model generalizes to search for targets efficiently in various context conditions, beats baseline models by a large margin, and demonstrates human-like visual search performance.

\section{Related Works} \label{sec:releatedworks}
Visual search in daily life is particularly challenging since the instance of the target object in the search scene could vary dramatically due to unexpected transformations like rotation, scale, color, occlusion, etc. Previously, a large body of behavioral\cite{wolfe2017five, miconi2015there, rao2002eye, rodriguez2007attention} and neurophysiological\cite{buschman2009serial, desimone1995neural, bichot2015source, sheinberg2001noticing} studies have focused on \emph{identical target} visual search, deviating largely from the complexities of real-world scenes. Recent works extend these efforts to establish human visual search benchmarks in approximately naturalistic conditions, where the search scenes were highly cluttered and the target information were provided as the text of the object category \cite{yang2020predicting} or an image instance different from the one in the search scene \cite{zhang2018finding}. 

\subsection{Bottom-up Saliency Prediction}
Certain parts of an image automatically attract attention due to color, contrast, or spatiotemporal changes. Several models have been proposed to capture such bottom-up saliency 
\cite{koch1987shifts,LIttiAmodel,JHarelGraph-based,LZhangSUN,NDBBruceSaliency,XHouImage,zhang2013saliency}, recently using deep convolution networks \cite{XunHuangSALICON,lin2014saliency, kruthiventi2017deepfix, kummerer2022deepgaze}. These models do not include information about the sought target.  

\subsection{Object Detection}
Many works\cite{eckstein2017humans, nicholson2022could} use object detection algorithms\cite{he2017mask} to perform visual search. Object detection models generate bounding boxes around several objects of interest. Generally, models for object detection include a convolutional neural network backbone with additional engineered components that use the output of the network to produce candidate bounding boxes. Compared to human visual search, object detection models employ brute-force strategies which are not driven by top-down cues. Critically, object detection algorithms are not able to generalize to search for \textit{novel} objects, but rather require extensive training with the target objects.

\subsection{Goal-directed Invariant Visual Search}
Humans rely heavily on the provided target as the task goal during visual search; such target-dependent modulation is likely originated from the prefrontal cortex\cite{bichot2015source, miller2001integrative} and projected onto lower-level visual cortical structures\cite{bisley2011neural, martinez2011searching}. There have been only a few attempts in using deep learning models to predict human fixation sequences in visual search tasks \cite{wei2016learned, adeli2018deep, zhang2018finding, yang2020predicting}, all of which were inspired by top-down modulation in humans. \cite{zhang2018finding} developed a convolutional neural network (CNN) that processes search and target image independently, where the attention map for visual search is computed by convolving the top-level search and target feature representation. While this model demonstrates high resemblance to human behavior in terms of search efficiency and scanpath history, it solely focuses on target-based modulation without considering other sources of information present in the search scene such as the surrounding context. Another work\cite{yang2020predicting} uses Inverse Reinforcement Learning (IRL) to learn human scanpaths by treating each fixation as a potential source of reward and trains a fixation sequence generator to maximize the reward. This model, however, requires supervised learning with human eye tracking data, which is labor-intensive and economically costly to obtain. Additionally, since the model was trained to perform the task using 18 pre-defined object categories, it could hardly generalize to all visual search conditions, such as novel categories outside of the training domain or objects without category labels. 

While CNNs have been widely adopted in computer vision algorithms, Transformer-based architectures -- particularly, the Vision Transformer (ViT)\cite{dosovitskiy2020image} -- has recently outperformed CNNs in object classification. 
Importantly, while ViT has not been constructed with explicit invariance properties like CNNs (with built-in translation invariance), it demonstrates stronger robustness against input perturbations including natural corruptions, real-world distribution shifts, and natural adversarial examples \cite{bhojanapalli2021understanding} when trained on sufficient amount of data. Such robustness against a broad range of perturbations is essential for invariant visual search.

The proposed TCT is more biologically plausible than other search models, as it integrates two essential properties that guide human searching behavior: target and context. These human-inspired attention modulators are incorporated into the pre-trained VIT architecture. 
Moreover, TCT does not require additional training or fine-tuning on human eye tracking data or ground truth labels for visual search tasks. 


\subsection{Contextual Modulation}
A few psychophysics experiments have been carried out to establish human benchmarks in in- and out-of-context object recognition and showed that incongruent contextual cues impair recognition performance \cite{zhang2020putting, bomatter2021pigs}. 
Computer vision models in object recognition can learn co-occurrence statistics between an object's label and visual appearance but also its label and surrounding context\cite{divvala2009empirical, sun2017seeing, beery2018recognition}. The multi-head self-attention blocks in ViT \cite{dosovitskiy2020image} compute long-range interactions between local image patches, by definition carrying statistical information between the object of interest and the surrounding context. A recent work\cite{bomatter2021pigs} introduced a Context-aware Recognition Transformer Network (CRTNet) that processes object and context information in two independent streams before integrating them in subsequent decoder modules. With an additional confidence estimator that balances the weights of the two streams, CRTNet could recognize out-of-context objects robustly. However, we still lack such computational models for visual search tasks.

The aforementioned IRL model \cite{yang2020predicting} attempts to integrate contextual information in visual search by including features of the target as well as other ``anchor objects'' \cite{vo2019reading, boettcher2018anchoring} and background scenes to model the internal belief state. The belief state representation is computed based on ``contextual beliefs'' of a limited number of object and scene categories extracted from a panoptic segmentation model\cite{kirillov2019panoptic, kirillov2019panoptic_1} and gets updated dynamically throughout the search process. The segmentation model, however, requires additional supervised training if the search task is to be generalized to novel object categories. Overall, such an approach assumes that humans have an internal scene parser that segments the search scene into object and scene classes. Our model, on the other hand, does not make any presumption upon the mechanism of the contextual reasoning process and does not require segmentation. Instead, TCT transfers the contextual knowledge of the search scene obtained implicitly during object recognition tasks to visual search tasks.

\section{Our TCT Model} \label{sec:model}
We propose a Target and Context-aware Transformer (TCT) to perform naturalistic visual search tasks with high efficiency. TCT takes in a target object $\mathbf{I}_T$ and a search scene $\mathbf{I}_S$ and extracts feature representation of the target and the context independently. The extracted features are then applied onto each individual transformer self-attention layer $l$ in the form of target modulation $\mathbf{M}_{T,l}$ and context modulation $\mathbf{M}_{C,l}$ to guide self-attention map $\mathbf{SA}_{l}$. We call each of these modulated self-attention layers a Target and Context-aware Attention Block (TCAB). The visual search task is performed on the attention map extracted from the final TCAB. 

\begin{figure*}[ht]
\begin{center}
\includegraphics[width=17.5cm]{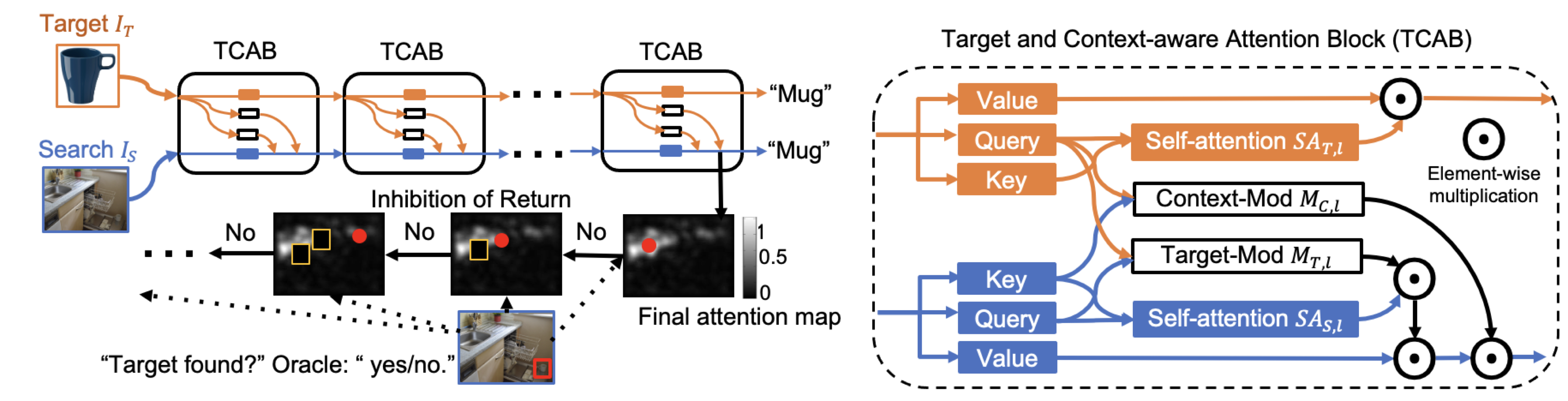}\vspace{-4mm}
\end{center}
   \caption{\textbf{Architecture for Target and Context-aware Transformer (TCT)}. TCT takes in a target object $\mathbf{I}_T$ and a search scene $\mathbf{I}_S$ and extracts feature representation of the target and the context independently. The extracted features are applied onto each Target and Context-aware Attention Block (TCAB) in the form of target modulation $\mathbf{M}_{T,l}$ and context modulation $\mathbf{M}_{C,l}$ to guide attention and in turn produce a final attention map. The model produces fixations by orderly selecting the maxima of the attention map.
   If the fixated area overlaps with the target bounding box, the search process ends. Otherwise, inhibition-of-return (IOR) is applied to the attention map by reducing the activation response within the fixated area to zero permanently in an area of pre-defined size. The process is repeated until the target is found. Red dots denote predicted eye fixations. 
   }\vspace{-5mm}
\label{fig:model}
\end{figure*}

We design the TCT architecture closely following that of the original Vision Transformer (ViT). This setup allows us to borrow the pre-trained ViT directly and easily re-purpose it into a visual search model 
without additional training. A schematic of TCT is shown in \textbf{Fig.~\ref{fig:model}}. Each of the self-attention layers in the original ViT is converted into a TCAB, while the remaining architecture stays completely intact and thus is not be elaborated here. 

\subsection{Feature Extraction for Target} \label{subsec:targetextraction}
We take the Vision Transformer(ViT) pre-trained on ImageNet-21K and fine-tuned on ImageNet-1K \cite{deng2009imagenet} with the official checkpoints from Google. We further fine-tune the ViT on CIFAR-100\cite{krizhevsky2009learning} while explicitly removing the position embeddings in order to exclude the spatial information in feature representation that is irrelevant for visual search tasks. 

The transformer reshapes an image of the target object $\mathbf{I}_T \in \mathbb{R}^{C\times W\times H}$ into a sequence of image patches $\mathbf{p}_T \in \mathbb{R}^{C\times N\times P^2}$, where $H$ and $W$ is the resolution of the original image, $C$ is the number of channels, $N$ is the number of patches given a pre-defined patch size $P\times P$. The patches are then converted into a sequence of patch embeddings $\mathbf{h}_{T,1} \in \mathbb{R}^{N\times D}$, where $D$ is the size of the hidden states in each transformer layer. In each layer $l$, the hidden states $\mathbf{h}_{T,l}$ are mapped through a linear projection $\mathbf{U}_{QKV, l}$ into query $\mathbf{Q}_{T,l}$, key $\mathbf{K}_{T,l}$, and value $\mathbf{V}_{T,l}$ representations, 
\begin{equation} \label{eqn:h2qkv}
    [\mathbf{Q}_{T,l}, \mathbf{K}_{T,l}, \mathbf{V}_{T,l}] = \mathbf{h}_T \mathbf{U}_{QKV,l}  
\end{equation}
each one of which is then reshaped into $N \times D_h$ for computing multihead self-attention, where $D_h$ is the hidden state size of each attention head. We retain the target query from each layer $\mathbf{Q}_{T,l}$ to represent the target feature and apply it in the corresponding layer to modulate attention when passing the search image $\mathbf{I}_S$ through the transformer (\textbf{Sec.~\ref{subsec:modulation}}).


\subsection{Feature Extraction for Context} \label{subsec:featureextractioncontextmod}
We extract contextual feature representations from a modified version of the CRTNet \cite{bomatter2021pigs}, an object classification transformer that integrates object information and contextual reasoning with cross-attention mechanisms. The original CRTNet architecture remains intact except that the confidence weighting system is pruned away, enforcing the network to learn more context information that benefits object recognition. For each pair of context image $\mathbf{I}_S$ (i.e. the search image in visual search settings) and target image $\mathbf{I}_T$, the modified CRTNet extracts context feature embeddings from a stack of transformer decoder layers. Subsequently, a context classifier, which consists of fully-connected layers and a softmax layer, takes the feature embedding from the last decoder layer and outputs the predicted class label probabilities $\mathbf{y}_{T, C}$ among $\mathbf{C}$ object classes. The modified CRTNet is trained end-to-end using cross-entropy loss with respect to $\mathbf{y}_{T, C}$.  

To further boost the contextual reasoning ability during object recognition, we introduced two changes for CRTNet training: \textbf{(1)}
   We 
   expand the COCO-18 training dataset (\textbf{Sec.~\ref{subsec: datasets}})
   with additional images from corresponding 18 object categories in MSCOCO\cite{mscoco}, generating a total of 201,383 target-context image pairs. 
   \textbf{(2)} The original CRTNet was trained on target-context image pairs where the target image was simply cropped out from the context image. We 
   perform 4 augmentations on the target-context pairs,
   combining 2 augmentations on target image $I_T$ and 2 augmentations on context image $I_S$. Augmentations on target images $I_T$ include: (1) the original target image $I_T$ is replaced with a random target image belonging to the same target category; (2) the original target image $I_T$ is blacked out with pixel value 0. Augmentations on context images $I_S$ include: (1) the target image $I_T$ within the original context image $I_S$ is cropped out and replaced with a random target belonging to the same category as $I_T$; and (2) the target image $I_T$ within the original context image $I_S$ is blacked out with pixel value 0.

To explicitly test whether the contextual feature extracted from the modified CRTNet contains useful information for object recognition, we check the top-1 recognition accuracy for the augmented target-context pairs where both the target itself and the target within the context images are blacked out (\textbf{Tab. S1}). Although the average accuracy for such inputs is much lower compared to the original intact target-context pairs (85\% vs 99\%), it is still significantly higher than chance level ($1/18=5.5\%$). This suggests that the surrounding context by itself carries statistical information that enables CRTNet to make reasonable inferences about the target object identity. Here we transfer such useful contextual information directly to our TCT model to aid visual search tasks.  



\subsection{TCTB} \label{subsec:modulation}
The transformer first takes in a search scene $\mathbf{I}_S$ and extracts its query $\mathbf{Q}_{S,l}$, key $\mathbf{K}_{S,l}$, and value $\mathbf{V}_{S,l}$ representations similarly as for the target image (\textbf{Sec.~\ref{subsec:targetextraction}}, \textbf{Eqn.~\ref{eqn:h2qkv}}). Within each TCAB, the unmodulated self-attention $\mathbf{SA}_l$ is computed as in the original ViT:
\begin{equation} \label{eqn:attn_self}
    \mathbf{SA}_l = \softmax\Bigl(\mathbf{Q}_{S,l} \mathbf{K}_{S,l}^\intercal / \sqrt{D_h}\Bigr).
\end{equation}

Intuitively, target modulation is computed as patch-wise local relevance between the target and search image, whereas contextual modulation is applied onto the attention output in a global fashion. For target modulation, we aim to locate the search patches that contain similar visual features as the target patches. This is implemented by first computing the cross attention between the target query $\mathbf{Q}_{T,l}$ and the search query $\mathbf{Q}_{S,l}$ and then voting for the most relevant search patch for each target patch $\mathbf{p}_T$. Voting is implemented as assigning the top one most relevant search patch with relevance $1$ and all others with relevance $0$ (denoted as $\keeptop(\cdot)$). 
\begin{equation}\label{eqn:mod_t}
    \mathbf{M}_{T,l} = \keeptop\biggl(\softmax\Bigl(\mathbf{Q}_{T,l} \mathbf{Q}_{S,l}^\intercal / \sqrt{D_h}\Bigr)\biggr)  
\end{equation}
The resultant binary target-search relevance map serves as our target-driven attention modulator $\mathbf{M}_{T,l}$, which is applied onto the original search image self-attention weights $\mathbf{SA}_l$ in an element-wise fashion (denoted as $\odot$) to achieve target-modulated attention weights $\mathbf{A}_{T,l}$.
\begin{equation}\label{eqn:attn_mod}
    \mathbf{A}_{T,l} = \mathbf{M}_{T,l} \odot \mathbf{SA}_l 
\end{equation}
The attention output computed based on target-modulated attention weights $\mathbf{A}_{T,l}$ and the search value $\mathbf{V}_{S,l}$ is then further modulated by the contextual modulator $\mathbf{M}_{C,l}$ (detailed in \textbf{Sec.~\ref{subsec:featureextractioncontextmod}}) to achieve the updated hidden states $\mathbf{h}_{S,l+1}$ that serve as the input for the next TCAB. 

 \begin{equation} \label{eqn:attn_output}
    \mathbf{h}_{S,l+1} = (\mathbf{M}_{C,l} + \mathbf{1}) \odot \left(\mathbf{h}_{S,l} + \mathbf{A}_{T,l} \mathbf{V}_{S,l} \right) 
\end{equation} 

 While both target and contextual modulation happen within each TCAB, their influence propagates through the entire multi-block hierarchy of the transformer. Importantly, the class token is exempt from the target and contextual modulation operations within each block but does inherit the modulation effects from previous blocks. We use the class token averaged across all attention heads from the last block as the final attention map to perform visual search tasks. 






\subsection{Fixation Sequence Generation} \label{subsec:IOR}
The ground truth location is defined as the bounding box of the target object, i.e., the smallest rectangle encompassing all pixels of the object. A fixation sequence is generated by iteratively selecting the maximal point in the attention map as the fixation center. For each fixation, we check whether the fixated area -- a rectangle of pre-defined size centered at the fixation center -- overlaps with the target bounding box. If so, the search process ends; otherwise, inhibition-of-return (IOR)\cite{klein2000inhibition} is applied to the attention map by reducing the activation response within the fixated area to zero. This reduction is permanent, in other words, infinite memory is assumed for inhibition of return. The search continues until the target is found. 
The oracle approach of judging whether the target has been located is a simplification; in the future, the model could include a recognition module to check whether that target can be detected at a certain fixation \cite{zhang2018finding}.


\subsection{Implementation Details and Code Availability}

Both target and contextual modulation in TCT are implemented on individual layers of the ViT and therefore it is easy to investigate the contribution of each type of modulation in different combinations of layer configurations (see layer ablation results in \textbf{Sec.~\ref{subsec:ablation}}). TCT achieves highest search efficiency when target modulation is implemented at $l\in [6, 12]$ and context modulation solely at $l=3$. The target or context modulator ($\mathbf{M}_{T,l}$ or $\mathbf{M}_{C,l}$) is set to a matrix of all ones $\mathbf{1}^{N_S\times N_S}$ or $\mathbf{1}^{N_S\times D}$, respectively, if no such modulation is implemented at the corresponding layer $l$, where ${N_S}$ is number of patches in $I_S$ and $D$ is the hidden state size.
We perform the visual search task using search image resolution and IOR size consistent with previous works that introduced the image datasets \cite{yang2020predicting, zhang2018finding}: $512 \times 320$ pixels as search image and $48 \times 48$ pixels as IOR for the COCO-18 and SCEGRAM datasets, and $1280 \times 1024$ as search image and $200 \times 200$ as IOR for the NatClutter dataset, respectively (\textbf{Sec.~\ref{subsec: datasets}}). 
All the source code is available at
\href{https://drive.google.com/drive/folders/1SgGrfhAhUNNjzxQ6cWMT7m9hm5i5QypN?usp=share_link}{here}.

\begin{figure*}[t]
\centering
\subfloat[COCO-18]{\includegraphics[width=5.2cm]{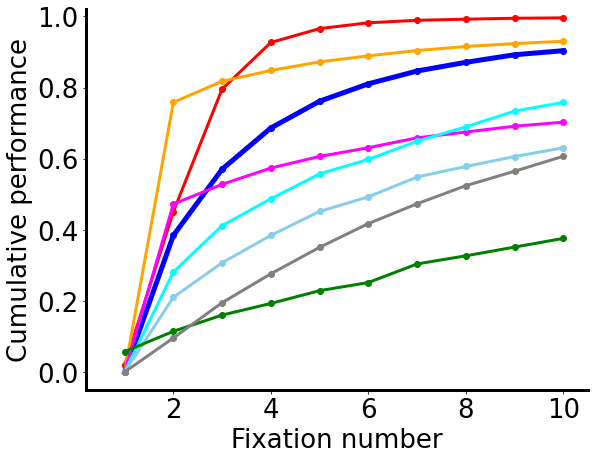}\label{fig:fig3a}}\hspace{-0.2cm}
\subfloat[NatClutter]{\includegraphics[width=5cm]{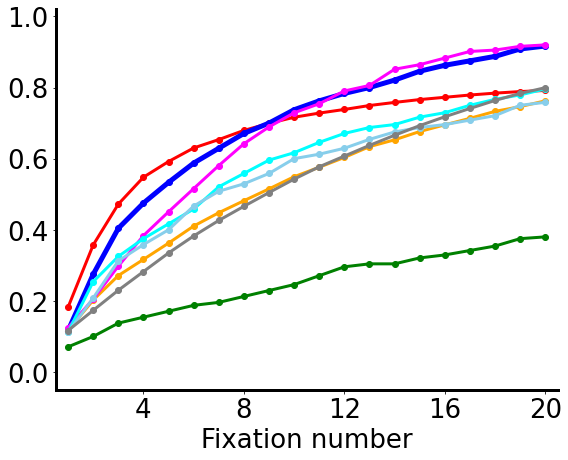}\label{fig:fig3b}}\hspace{-0.2cm}
\subfloat[SEAGRAM]{\includegraphics[width=5cm]{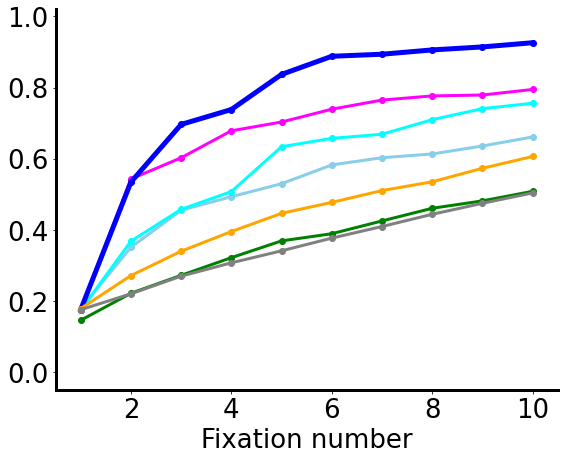}\label{fig:fig3c}}\hspace{-0.2cm}
\subfloat{\includegraphics[width=2.5cm]{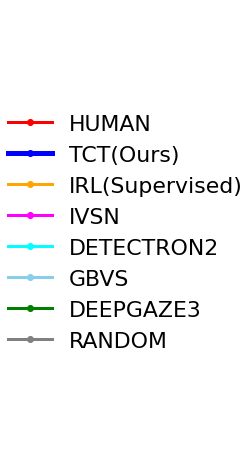}}\vspace{-2mm}
   \caption{\textbf{Cumulative Performance of TCT and other baseline models on COCO-18 \textbf{(a)}, NatClutter \textbf{(b)}, SEAGRAM \textbf{(c)}}. Cumulative search performance is calculated as the cumulative probability $p(n)$ to find the target in $\le n$ fixations for humans (red), TCT (blue) and other competitive methods (see \textbf{Sec}.~\ref{subsec: baselines}). Human fixation data is not available in the SCEGRAM dataset.}
\label{fig:fig3}\vspace{-4mm}
\end{figure*}

\section{Experiments} \label{sec:experiment}
\subsection{Datasets} \label{subsec: datasets}


\begin{table}
  \centering
  \footnotesize
  \begin{tabular}{lllll}
    \toprule
    Attributes & \# Subjects & \# Trials & Target Ratio & Incongruency\\
    \midrule
    COCO-18\cite{yang2020predicting}    & 10    & 612    & 0.04      & No \\
    NatClutter\cite{zhang2018finding}   & 15    & 240    & 0.01      & No \\
    SCEGRAM\cite{scegram}               & N/A   & 187    & 0.01      & Yes \\
    \bottomrule
  \end{tabular}
  \caption{\textbf{Dataset summary.} Characteristics of the three image datasets used for performing visual search tasks: number of human subject with eye tracking data (``\# Subjects"), number of search trials (``\# Trials"), average ratio between the target and search image size (``Target ratio"), existence of search scenes with incongruent contexts (``Incongruency").  
  }
  \label{tab:dataset}\vspace{-4mm}
\end{table}

\textbf{COCO-18}\cite{chen2021coco} is a visual search dataset consisting of about 300,000 eye movements from 10 human subjects in searching for one of 18 target object categories in 6,202 natural indoor scenes from MSCOCO dataset \cite{mscoco}. The target object is congruent with the context in each natural image. As the target images are not provided in the original dataset, 
we paire each search image
with a randomly selected target image from the same object category as the one present in the original search image.

\textbf{NatClutter}\cite{zhang2018finding} consists of 240 invariant target-search image pairs in complex scenes with congruent contexts. The sought target and the given target differ in geometric transformations. The target object category could be novel to all models of interest in the current study.

\textbf{SCEGRAM}\cite{scegram} consists of 62 indoor target objects, each contextualized in 1 congruent search scene and 5 additional scenes with semantic and/or syntactic violations. The target rendered within the search images and the provided target image originate from the identical objects. To enable invariant visual search, we collect a novel set of 62 target objects varying in appearances as well as geometric transformations from the sought targets in the search images. Additionally, since context modulation is more important for smaller objects \cite{zhang2020putting,bomatter2021pigs}, we sort the search images according to the ratio of target versus search image size in ascending order and only consider the bottom 50 percentile in the current study. In the modified SCEGRAM dataset, we have 187 images in total, with 32 congruent conditions and 155 incongruent conditions.


\subsection{Baselines and metrics}\label{subsec: baselines}
\textbf{IVSN}\cite{zhang2018finding} is a zero-shot visual search CNN with top-down target-driven modulation and outputs an attention map in each search trial.
\textbf{IRL}\cite{yang2020predicting}. 
is an inverse reinforcement learning model trained on human fixation sequences on COCO-18 to predict visual search scanpaths.
When using IRL to perform search on objects without explicit category labels such as in NatClutter and SCEGRAM,
we iterate through the 18 training categories to compute a category-average attention map.
\textbf{Detectron2}\cite{detectron2} has a collection of object detection algorithms.
We used 
ResNet+FPN pre-trained on MSCOCO \cite{mscoco} as the backbone for visual search. 
The attention map is computed as the average of sigmoid objectness maps extracted from the RPN component.
\textbf{GBVS} (Graph-based Visual Saliency)\cite{gbvs} is a bottom-up saliency model predicting attention maps based on low-level statistics of the search images.
\textbf{Deepgaze\uppercase\expandafter{\romannumeral3}}\cite{kummerer2022deepgaze} is a bottom-up saliency model trained with human eye fixations in free-viewing tasks and outputs bottom-up attention maps for search images.
We also consider a \textbf{Random} model that populates attention maps with values randomly drawn from a uniform distribution. This randomly sampling process is repeated 100 times. 

For all the baseline models above, we generate a sequence of fixations by iterating through the maxima of the attention maps while applying IOR throughout the search process (see \textbf{Sec}.~\ref{subsec:IOR}). We evaluate all computational models with the following evaluation metrics:
\noindent  \textbf{Cumulative performance across fixations} is calculated as the cumulative probability $p(n)$ that a human subjects or a computational model finds the target object in $\le n$ fixations. 
We also compute \textbf{Average number of fixations} until the target object has been found.


\section{Results} \label{sec:restuls}
\begin{figure*}[ht]
\begin{center}
\includegraphics[width=17.5cm]{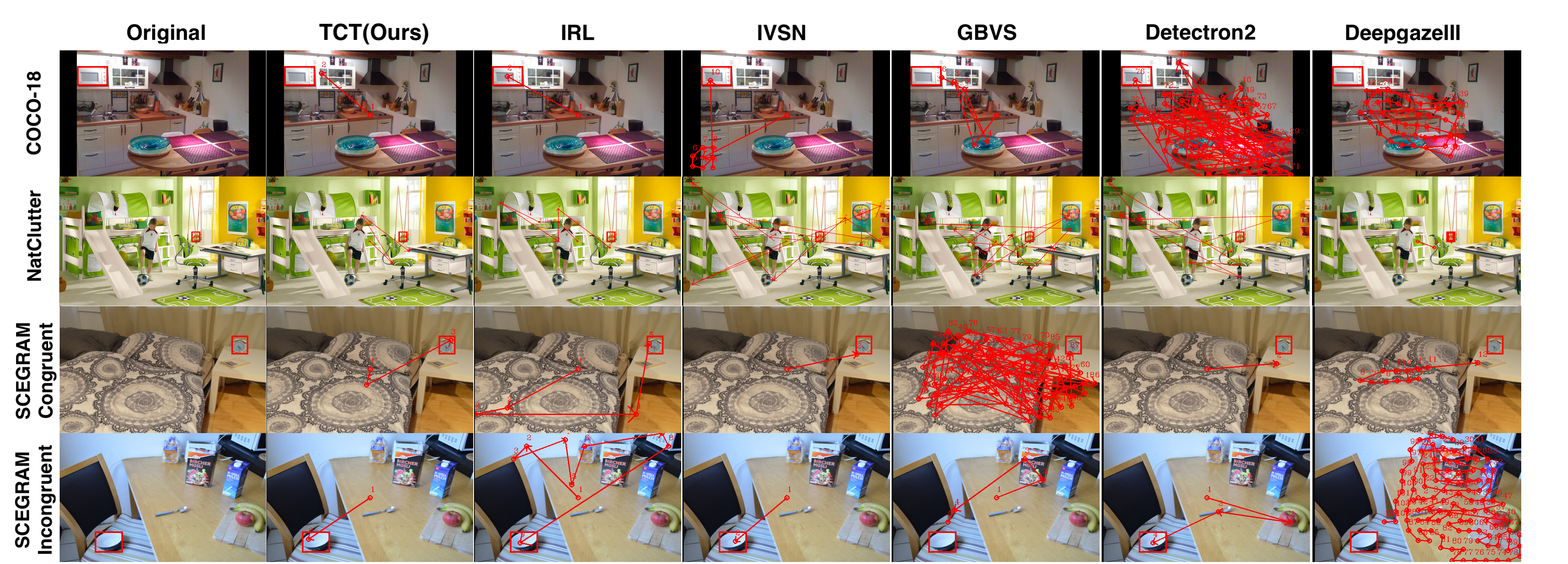}\vspace{-4mm}
\end{center}
   \caption{\textbf{TCT performs visual search tasks efficiently in multiple datasets and across different context conditions.} Visual search scanpaths of TCT (column 2) and other baseline models (column 3 to 7) on COCO-18 (row 1), NatClutter (row 2), SCEGRAM Congruent (row 3) and SCEGRAM Incongruent (row 4) datasets. The red bounding box shows the target object. Red dots denote predicted eye fixations. Red arrows indicate predicted eye movement directions.
   }\vspace{-5mm}
\label{fig:scanpath}
\end{figure*}

\subsection{Visual Search on Natural Images}


We first evaluate the visual search performance of TCT against the human benchmark as well as other baseline visual search models on natural images in the COCO-18 dataset (\textbf{Fig.~\ref{fig:fig3a}}). As shown in the example scanpaths in \textbf{Fig.~\ref{fig:scanpath}} row 1, TCT is able to locate the target object rapidly despite the object is far away from the initial fixation position (the center of the image). TCT manages to find the object using only 1 fixation for 38\% of all trials and no more than 10 fixations for 92\% of all trials. 

While TCT behaves worse than human in the initial fixations, its cumulative performance increases rapidly across fixations and reaches close to human performance within 10 fixations. IRL serves as another upper bound performance for this task, since it was trained specifically on this dataset using human fixation sequence to mimic human search behavior. 

Among all of the zero-shot visual search models,
our TCT achieves the highest search efficiency (\textbf{Fig~\ref{fig:fig3a}}). In the example kitchen scene, TCT efficiently locates the microwave with 1 fixation (\textbf{Fig~\ref{fig:scanpath}}), while other baseline models took more fixations to find the target. 
The outstanding performance can be most likely attributed to its human-inspired integration of both target and context information in guiding search attention. In contrast to TCT, IVSN is ignorant of any contextual cues and solely relies on target information to generate attention, leading to lower search efficiency. TCT is also notably better than Detectron2 since the latter utilizes object detection algorithms and is not aware of top-down modulation from the object of interest. GBVS and Deepgaze\uppercase\expandafter{\romannumeral3} also exhibit evidently lower performance, both of which are bottom-up saliency models that do not take the target object information into account. 

\subsection{Visual Search in Cluttered Environments}


To evaluate visual search performance on more complex scenes where the target objects are smaller in size and possibly occluded by multiple distractors, we test TCT and other baseline models on the NatClutter dataset. Example scanpaths in row 2 of \textbf{Fig~\ref{fig:scanpath}} demonstrate a challenging search trial with a small teddy bear embedded in a highly cluttered scene with several distractors.  Although the target object appears much less salient than in COCO-18, our TCT is still able to pinpoint the target position within a few fixations. TCT successfully locates the object using only 1 fixation for 28\% of all trials and no more than 20 fixations for 92\% of all trials.  

While human subjects outperform TCT in the first 8 fixations, TCT surpasses human in subsequent fixations. One reason for this difference is that humans do not have infinite memory and potentially revisit the same fixation areas \cite{zhang2022returnfixations}, whereas permanent IOR has been applied in all computational models, avoiding exhaustive search with repeated visits. Meanwhile, The cumulative search performance of IRL dropped dramatically in NatClutter compared to COCO-18, since IRL has been implemented to perform the search only within its limited training categories within COCO-18 and therefore fails to generalize to novel object categories or unlabelled objects. TCT manages to outperform IRL by a large margin due to its ability to generalize to novel target objects with minimal or even zero prior exposure.

In general, TCT performed better than other baseline models. IVSN, particularly, achieves performance similar to TCT. Both TCT and IVSN use target modulation and generalize well to novel target objects. Nonetheless, the cumulative performance of IVSN was still lower than TCT possibly due to the lack of context modulation.

\subsection{Visual Search in Incongruent Contexts}
\begin{table}
  \centering
  \footnotesize
  \begin{tabular}{lllll}
    \toprule
    Avg Fixations & Overall & Congruent & Incongruent \\
    \midrule
    TCT(ours)    & \textbf{4.2}& \textbf{3.6} & \textbf{4.9} \\
    IRL           & 12.8    & 8.9       & 16.8 \\
    IVSN          & 6.6     & 5.2       & 8.1\\
    GBVS          & 15.3    & 10.7      & 19.8 \\
    Detectron2    & 11.7    & 5.6       & 17.8 \\
    Deepgaze3     & 24.1    & 18.3      & 29.8\\
    \bottomrule
  \end{tabular}
  \caption{\textbf{Average number of fixations for TCT and baseline models under congruent and incongruent conditions in the modified SCEGRAM dataset}. Best is in bold.}\vspace{-4mm}
  \label{tab:con_incon_result}
\end{table}

So far, we have focused on searching in natural scenes with contextual cues congruent with the target objects. Next, we assess whether target-context incongruency diminishes the search efficiency of visual search models using the modified SCEGRAM dataset (\textbf{Sec}.~\ref{subsec: datasets}). Search trials in SCEGRAM are split into congruent and incongruent conditions and performance of each model is computed for the two different conditions separately (\textbf{Table}~\ref{tab:con_incon_result}). As demonstrated in the example scanpaths (\textbf{Fig.~\ref{fig:scanpath}}, rows 3-4), both TCT and IVSN exhibit small performance difference across congruent and incongruent conditions, whereas other baseline models suffer severely from incongruent contexts. These results agree with our expectation because both TCT and IVSN modulate attention with target information and therefore are somewhat protected against misleading contextual information, enabling efficient search even under incongruent conditions.

\begin{figure}[t]
\begin{center}
\includegraphics[width=7cm]{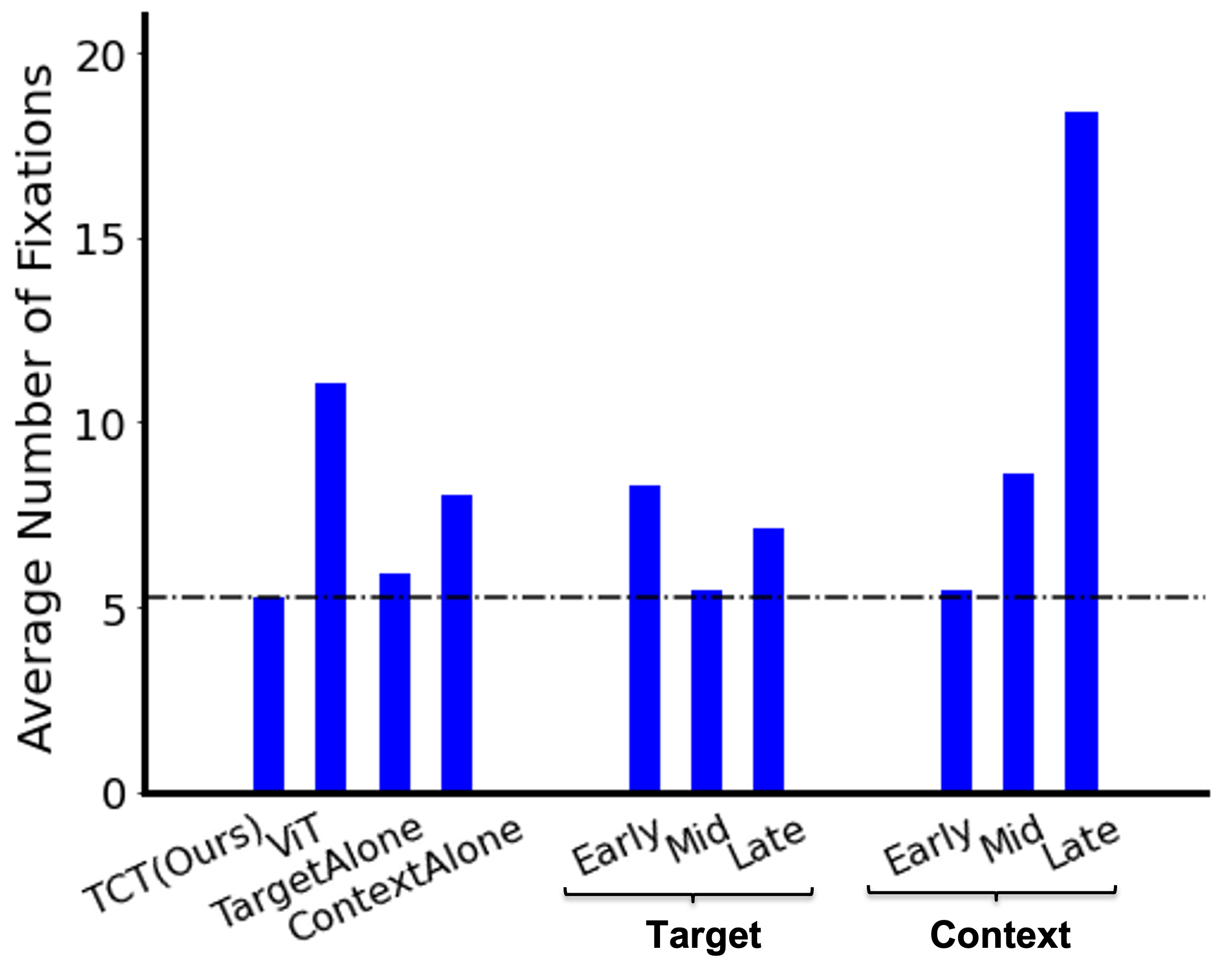}\vspace{-4mm}
\end{center}
   \caption{\textbf{COCO-18 search performance for TCT with ablated components or with modulation applied on different layers}. Performance is evaluated as the average number of fixations and the dash line indicates the best TCT performance out of all different ablations of TCT. Left 4 columns: our default TCT, TCT with target modulation alone (TargetAlone), TCT with context modulation alone (ContextAlone), the original ViT with no modulation (ViT). Mid 3 columns: TCT with target modulation applied starting from early, mid, and late TCAB blocks onwards,  respectively. Right 3 columns: our TCT with context modulation applied at individual early, mid, and late TCAB blocks, respectively. 
   }\vspace{-5mm}
\label{fig:ablation}
\end{figure}

\subsection{Ablation}\label{subsec:ablation}


To further dissect individual contribution of target and contextual modulation on search efficiency, we repeat the identical experiments on COCO-18 using different ablated versions of TCT and compute their average number of fixations (\textbf{Fig.~\ref{fig:ablation}} left). When both target and context modulation are removed, TCT becomes the original ViT (``ViT") and performs poorly as it simply functions as a bottom-up feature extractor and self-attention does not contain information relevant for the visual search task. To evaluate the importance of context modulation, we remove the global context modulation while keeping the local target modulation intact (``TargetAlone''). The TargetAlone model is still able to provide goal-directed information in a top-down fashion similar to its CNN counterpart IVSN \cite{zhang2018finding}, yet the performance drops in comparison to TCT due to lack of the contextual guidance. Similarly, we remove the local target modulation while leaving the global context modulation intact (``ContextAlone''). The ContextAlone model performs significantly worse than TCT, indicating that target-driven modulation is essential in guiding search attention. While context information guides attention on a global scale to emphasize regions of interest for contextual reasoning given a target object, by itself it does not provide information that is high resolution enough to pinpoint the object location. 

We further investigate the layer specificity of target and contextual modulation by applying the modulation on different subsets of TCABs for each modulation type independently. The target modulation is applied starting from a certain block and continues across subsequent blocks. We group the 
blocks where the modulation happens into early ($l \in [1, 4]$), middle ($l \in [5, 8]$), and late ($l \in [9, 12]$) groups. For each group, we average the search performance. The best performance is achieved when target modulation is applied from the middle blocks onward (\textbf{Fig.~\ref{fig:ablation}} middle), potentially because the early blocks in transformer mainly extract low-level nuisance visual properties such as contrast or color that do not serve as essential cues in visual search. Similarly, we apply the contextual modulation onto different individual layers and group the modulation blocks in the same fashion. TCT with contextual modulation in earlier blocks significantly outperforms than that in later layers (\textbf{Fig.~\ref{fig:ablation}} right), aligning with the intuition that context functions as global guidance that redirects attention at the initial stage of visual processing. 

\section{Conclusion} \label{sec:conclusion}
We propose a biologically plausible deep learning architecture for visual search, \textbf{TCT} (Target and Context-aware Transformer). TCT integrates target object and contextual information via a multi-head transformer encoder to guide attention during visual search tasks. TCT transfers contextual knowledge learnt from object recognition to visual search in a zero-shot manner. TCT manages to approximate human search efficiency in invariant zero-shot visual search tasks with various levels of difficulty. Remarkably, in contrast with standard object detection approaches, TCT generalizes across tasks with novel target object search \emph{without retraining or fine-tuning}. Furthermore, TCT outperforms the SOTA visual search models, especially in highly complex datasets and across congruent and incongruent contexts in a newly introduced dataset.






\onecolumn
\newpage
\clearpage

\renewcommand{\thesection}{S\arabic{section}}
\renewcommand{\thefigure}{S\arabic{figure}}
\renewcommand{\thetable}{S\arabic{table}}
\setcounter{figure}{0}
\setcounter{section}{0}
\setcounter{table}{0}


\section*{Supplementary Materials}

\begin{table}[ht]
  \centering
  \footnotesize
  \begin{tabular}{@{}lc@{}}
    \toprule
                                         & \textbf{Top-1 Accuracy(\%)}  \\
    \midrule
    Original Context, Original Target    & 98.8     \\
    Original Context, Random Target      & 98.6     \\
    Zero Context, Zero Target            & 84.9     \\
    \bottomrule
  \end{tabular}
  \caption{\textbf{Top-1 object recognition accuracy of the context feature extractor on COCO-18}. Classification accuracy of the context feature extractor (see \textbf{Sec. 3.2}) under different context conditions
  in COCO-18 test dataset: (1) the target image $I_T$ is cropped out directly from the context image $I_S$ and the context image remains intact (``Original context, Original target"); (2) the original target image $I_T$ is replaced with a randomly selected target image belonging to the same target category as the original $I_T$,
  while the context image $I_S$ remains intact (``Original context, Random target");  (3) both the original target image $I_T$ and the bounding box encompassing the target region on the original context image $I_S$ is blacked out with pixel value 0 (``Zero context, Zero target").}
  \label{tab:top1accuracy}
\end{table}


\newpage
{\small
\bibliographystyle{ieee_fullname}
\bibliography{egbib}
}

\end{document}